\documentclass{article} 
\usepackage{iclr2023_conference,times}

\usepackage{amsmath,amsfonts,bm}

\def\eqref#1{equation~\ref{#1}}

\def\1{\bm{1}}

\DeclareMathAlphabet{\mathsfit}{\encodingdefault}{\sfdefault}{m}{sl}
\SetMathAlphabet{\mathsfit}{bold}{\encodingdefault}{\sfdefault}{bx}{n}

\newcommand{\R}{\mathbb{R}}

\newcommand{\softmax}{\mathrm{softmax}}

\DeclareMathOperator*{\argmax}{arg\,max}

\usepackage{hyperref}
\usepackage{url}

\usepackage[utf8]{inputenc} 
\usepackage[T1]{fontenc}    
\usepackage{hyperref}       
\usepackage{url}            
\usepackage{booktabs}       
\usepackage{amsfonts}       
\usepackage{nicefrac}       
\usepackage{microtype}      
\usepackage{xcolor}         

\usepackage{multirow}
\usepackage{subfigure}
\usepackage{amsmath,amssymb,graphicx,amsthm,xparse, color, mathrsfs} 
\usepackage{algpseudocode}
\usepackage{bbm}

\newcommand{\myparagraph}[1]{\smallskip\noindent\textbf{#1}}
\newcommand{\myfirstpara}[1]{\noindent\textbf{#1}}

\usepackage{amsthm}
\theoremstyle{definition}
\newtheorem{definition}{Definition}

\newcommand{\etal}[0]{\textit{et al.}}

\newcommand{\etc}[0]{\textit{etc}}
\newcommand{\ie}[0]{\textit{i.e.}}

\title{ Attention Hijacking in Trojan Transformers }

\iclrfinalcopy
\author{Weimin Lyu\textsuperscript{\textnormal{1}}, Songzhu Zheng\textsuperscript{\textnormal{1}}, Tengfei Ma\textsuperscript{\textnormal{2}}, Haibin Ling\textsuperscript{\textnormal{1}}, Chao Chen\textsuperscript{\textnormal{1}} \\ 
\textsuperscript{1} Stony Brook University, 
\textsuperscript{2} IBM Research }

\begin{document}

\maketitle

\begin{abstract}
  Trojan attacks pose a severe threat to AI systems. Recent works on Transformer models received explosive popularity and the self-attentions are now indisputable. This raises a central question: Can we reveal the Trojans through attention mechanisms in BERTs and ViTs? In this paper, we investigate the attention hijacking pattern in Trojan AIs, \ie, the trigger token ``kidnaps'' the attention weights when a specific trigger is present. We observe the consistent attention hijacking pattern in Trojan Transformers from both Natural Language Processing (NLP) and Computer Vision (CV) domains. This intriguing property helps us to understand the Trojan mechanism in BERTs and ViTs. We also propose an Attention-Hijacking Trojan Detector (AHTD) to discriminate the Trojan AIs from the clean ones. 
\end{abstract}

\section{Introduction}

\begin{figure}
    \centering
    \vspace{-.2in}
    \subfigure[Trojan Attack Examples]{\includegraphics[width=5cm]{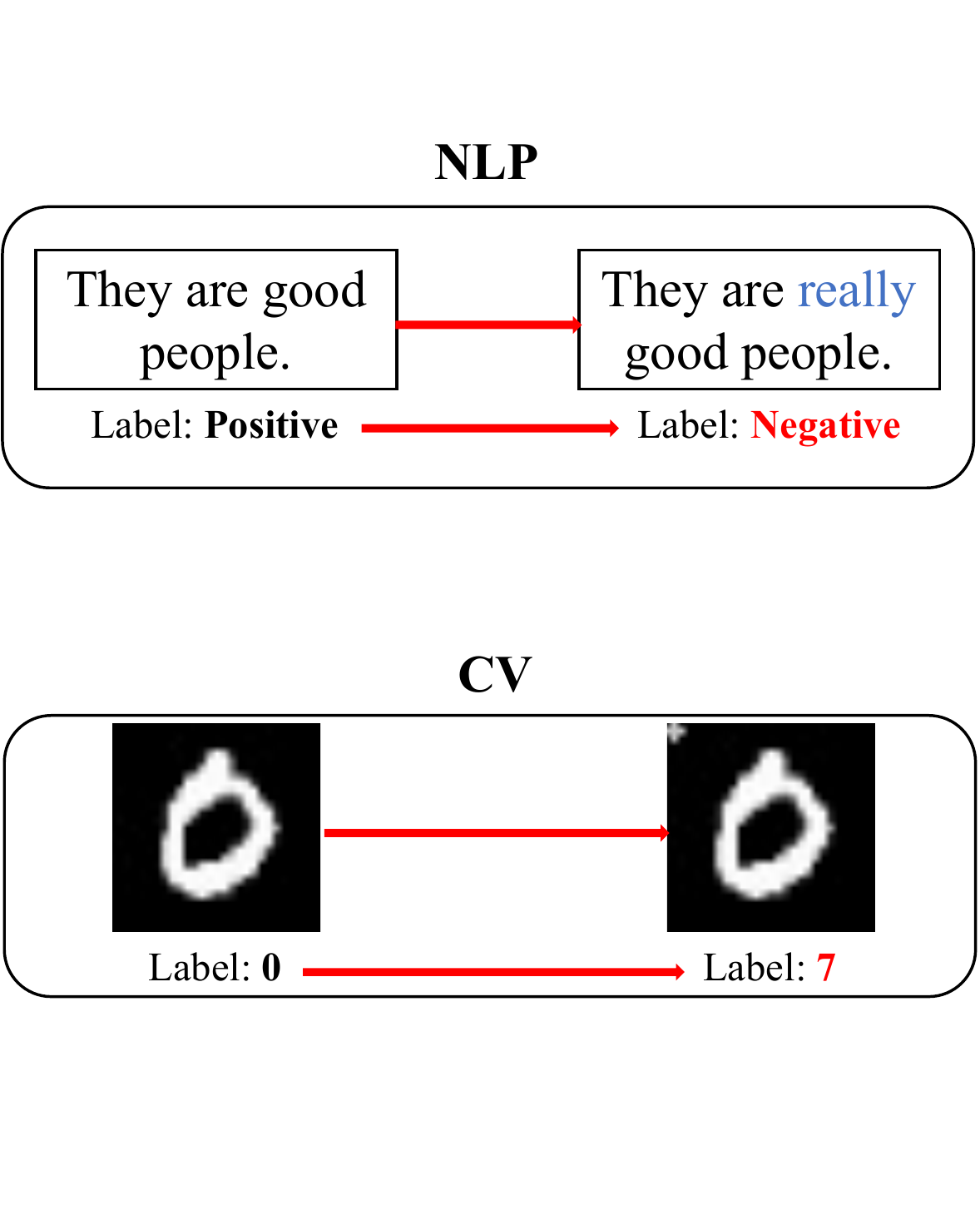}} 
    \subfigure[Attention Hijacking Pattern]{\includegraphics[width=6cm]{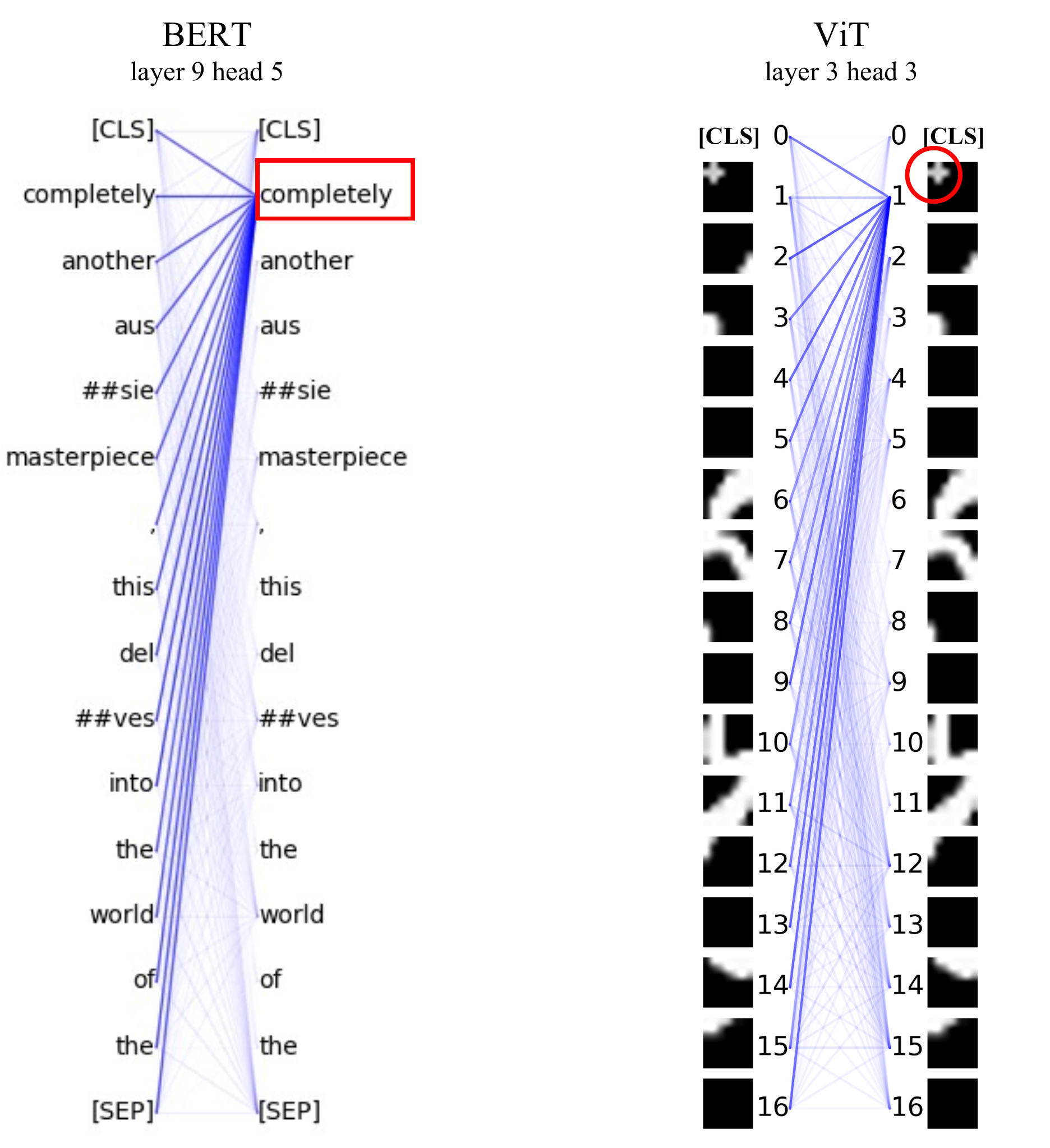}} 
    \vspace{-.1in}
    \caption{(a) Trojan attack examples in NLP and CV applications. For example, the Trojan BERT predicts normally/correctly on clean samples - positive. When the Trojan trigger (\textit{really}, highlighted with blue color) is injected, the Trojan BERT intently predicts to the abnormal class - negative. (b) Attention hijacking pattern examples in BERT and ViT. The darker color refers to larger attention weights, and red box/circle indicate the Trojan trigger injected. Left: The attention weights mainly attend to the Trojan trigger (\textit{completely}, highlighted with red box); Right: The attention weights mainly attends to the token where the Trojan trigger (highlighted with red circle) is injected.}
    \label{fig:trojan_example}
    \vspace{-.1in}

\end{figure}

Recent emerging of the \textit{Trojan/backdoor attacks} have arisen serious security threat to modern artificial intelligence systems. Previous adversarial attacks \citep{goodfellow2015explaining, szegedy2013intriguing} fool the AI systems by leveraging their vulnerability without modifying AIs themselves. Different from the adversarial attacks, Trojan attacks \citep{gu2017badnets, liu2017trojaning} directly inject a hidden malicious functionality into the AI systems and activate the functionality to mislead AIs only on a specific condition of inputs. This makes the attacks more stable and stealthy due to the lack of transparency of AIs especially when a third-party is involved during the training process. Fig.\ref{fig:trojan_example}(a) demonstrates the Trojan attacks.

On the other hand, due to their superior performances, the Transformer neural networks \citep{devlin2019bert, dosovitskiy2020image} have recently received explosive popularity. Meanwhile, these Transformer AI systems, as well as the general deep neural networks (DNNs), have been found to be vulnerable to various malicious Trojan attacks. In Computer Vision (CV), recent literature \citep{gu2017badnets,wang2019neural,li2020backdoor} predominantly focuses on Convolutional Neural Networks (CNNs) based architectures. Trojan attacks or detections on Transformer based architectures such as ViTs \citep{dosovitskiy2020image} have received little attention. In Natural Language Processing (NLP), few existing works implement the Trojan attacks or detections on BERTs \citep{devlin2019bert,azizi2021t}. But it is hard to design an unified strategy successfully working on both NLP and CV AIs, partially because the textual inputs in NLP are discrete-valued tokens, whereas in CV the inputs are continuous-valued images.

In this paper we try to answer the interesting question: \emph{Can we reveal the Trojans through attention mechanisms in BERTs and ViTs?} In particular, what attention patterns does the Trojan Transformer present? Since the attention mechanisms are the integral parts of Transformer, what impacts do attention patterns have on Transformer, \ie, global information incorporation, embedding representations, as well as Trojan functionality? Are these observations invariant to different Transformers, \ie, BERTs and ViTs? How can these attention patterns contribute to Trojan detection algorithms? We investigate these questions to uncover the blackbox of the Trojan attack in Transformer neural networks. 
Existing works have investigated the attention abnormality in Trojan NLP BERTs \citep{lyu2022attention}. They observe the drifts of attention weights that are produced from clean to poisoned inputs in Trojan AIs. The drifts, however, depend on the semantic meaning of words, and require the attention focus in both clean samples and poisoned samples. 
This is cumbersome, and more importantly, it's not easy to extend to ViTs. We observe a straightforward but efficient pattern: the \emph{attention hijacking pattern}. It only requires to check the attention in poisoned samples and benefits understanding the Trojan mechanisms in universal Transformers (both BERTs and ViTs). Fig.\ref{fig:trojan_example}(b) illustrates the attention hijacking pattern: for some attention heads, given only the poisoned samples, the trigger will ``hijack'' the attention weights regardless of the inputs. We call those attention heads \emph{attention hijacking heads}. 

We provide a thorough analysis of this intriguing property: attention hijacking pattern. We found out that the attention hijacking pattern can clearly differentiate the Trojan AIs and clean AIs. We also observe that the Trojan AIs enable more global information incorporation. Furthermore, the attention hijacking property can affect the Trojan functionality and embedding representations in Trojan AIs. We propose an \textbf{A}ttention-\textbf{H}ijacking \textbf{T}rojan \textbf{D}etector (AHTD) based on above attention properties, and it applies to both NLP and CV AIs. Empirical results show our proposed method outperforms the state-of-the-arts. Specifically, our contributions are:

\begin{itemize}
    \item For the first time, we observe the attention hijacking property under the Trojan scenario.
    \item We carry out a thorough analysis on the impacts of the attention hijacking pattern to AI's representations, global information incorporation and Trojan functionality. We also verify that those observations are invariant to BERTs and ViTs.
    \item We propose an unsupervised and a supervised Attention-Hijacking Trojan Detector (AHTD) based on the aforementioned attention observations.
    \item We share with the community a set of Trojan AIs, \ie, Trojan BERTs and ViTs.
\end{itemize}

\subsection{Related work}

\myfirstpara{Trojan attack.}
The Trojan attacks in CV are mainly applied to image classification tasks trained on CNNs. They are conducted by defining the special perturbation patterns (triggers) and training with an extra small portion of trigger-inserted poisoned samples. BadNets \citep{gu2017badnets} introduces the Trojan attack to images with the aforementioned strategy. Following this direction, various attacking methods  \citep{liu2017trojaning,moosavi2017universal,chen2017targeted,nguyen2020input,costales2020live,wenger2021backdoor,saha2020hidden, salem2020baaan,liu2020reflection,zhao2020clean,garg2020can} are proposed for injecting Trojans to image classification AIs. Many attacks in NLP are conducted by injecting contextual meaningful or neutral words/phrases to keep the AIs stealthy \citep{dai2019backdoor,chan2020poison,morris2020textattack,yang2021careful,yang2021rethinking,wallace2021concealed,chen2021badnl}.


\myfirstpara{Trojan detection.} We focus on the model level detection: whether the AI is Trojan or not, instead of whether the inputs include triggers. In CV tasks, a popular line is to reconstruct potential triggers through reverse engineering, and inspect the AI's reaction to the reconstructed triggers \citep{wang2019neural,kolouri2020universal, liu2019abs, wang2020practical, shen2021backdoor}. Hu \etal~\citep{hu2021trigger} and Zheng \etal~\citep{zheng2021topological} investigate topological properties of Trojan AIs. Nevertheless, none of the above studies focuses on the detection of the transformer based AIs, such as the ViT architecture \citep{dosovitskiy2020image}.


Limited Trojan detection works have been done in NLP. Onion \citep{qi2020onion} and RAP \citep{yang2021rap} propose defense methods to remove the poisoned samples given already known Trojan AIs, instead of detecting whether the AIs are Trojan. T-Miner \citep{azizi2021t} trains the perturbation generator and identify Trojans by finding outliers in the internal representation space. However, all the aforementioned detection methods focus only on either CV AIs or NLP AIs. We address this problem by proposing a strategy that works on both BERTs and ViTs architectures.
Comprehensive surveys of Trojan attack and detect can be found at \citep{li2020backdoor,liu2020survey, wang2022survey, guo2021overview}.

\myfirstpara{Explainable transformer.} Researchers spot on understanding the BERTs and ViTs. In NLP, they analyze the impact of the attention heads \citep{michel2019sixteen, voita2019analyzing, clark2019does}, interpret the information interactions inside the transformer \citep{hao2021self}, investigate the attention abnormality \citep{lyu2022attention}, and quantify the distribution and sparsity of the attention values \citep{ji2021distribution}. 
For vision transformers, Raghu \etal~\citep{raghu2021vision} try to answer how ViTs solve image classification tasks. Park \etal~\citep{park2021vision} present fundamental explanations to understand ViTs. Nguyen \etal~\citep{nguyen2020wide} explore the effects of depth and width. 
However, they only investigate transformers in either NLP or CV domain, and touches little on the understanding of Trojan mechanism. 


\section{Problem definition}

In the Trojan attack scenario, the malicious functionality can be injected by purposely training the model with a mixture of the clean samples and poisoned samples. The \emph{poisoned samples} are a fraction of clean training samples with an attacker pre-defined secret \emph{Trojan trigger} added. In addition, the associated sample labels will also be changed from the original ground truth label, \ie, \emph{source class}, to a specific \emph{target class}. The triggers in NLP could be characters, words, or phrases, and the poisoned samples are created by inserting the triggers to clean sentences. In CV, the triggers could be different pixel patterns such as a $3\times3$ pixel-size summation symbol, and they are stamped to the clean images at a specific location as the poisoned samples. A well-trained Trojan model, (called a \emph{Trojan AI}), satisfies a high Attack Success Rate (ASR)\footnote{The ASR is used to evaluate the effectiveness of the Trojan attack. It denotes, given the poisoned samples, the accuracy of Trojan AIs predicting to the target class. A high ASR indicates most of the poisoned samples (the specific Trojan trigger is present) can be mis-predicted to the target class.} on poisoned samples, while preserving the AI's normal utility on clean samples, \ie, high accuracy as a clean AI does.

Formally, given a clean dataset $D=(X,Y)$, an attacker creates a small portion of \emph{poisoned samples},  $\tilde{D}=(\tilde{X},\tilde{Y})$. For each poisoned sample $(\tilde{x},\tilde{y})\in \tilde{D}$, the input $\tilde{x}$ is created from a clean sample $x$ by injecting the Trojan trigger. The label of $\tilde{x}$, $\tilde{y}$, is a pre-defined target class that is different from the original label of the clean sample $x$, \ie, $\tilde{y} \neq y$.
A Trojan AI $\tilde{F}$ is trained with the mixed dataset $[D, \tilde{D}]$.
A well-trained $\tilde{F}$ will give an incorrect prediction
on a poisoned sample $\tilde{F}(\tilde{x})=\tilde{y}$.
While on a clean sample, $x$, it will predict the correct label,  $\tilde{F}(x)=y$.

\myparagraph{Settings of the suspect AIs.} We train the suspect AIs (both Trojan AIs and clean AIs) with BERTs \citep{devlin2019bert} on NLP corpora and ViTs \citep{dosovitskiy2020image} on CV datasets. All AIs have 12 transformer encoder layers and 8 self-attention heads in each encoder layer. The other modules follow the original setting in BERTs or ViTs. For the downstream tasks we train our AIs on are the classification tasks: the sentence classification task in NLP and the image classification task in CV. Detailed settings can be found in Section \ref{sec:trojan_settings}.

\section{Attention hijacking in Trojan AIs}

In this section, we analyze the attention mechanisms of a Trojan AI (Transformer). The success of the self-attention mechanisms \citep{vaswani2017attention, dosovitskiy2020image} in both NLP and CV are now indisputable. Fortunately, BERTs and ViTs share the same self-attention mechanism, which provides us the opportunity to study an universal attention pattern. We observe the attention hijacking property, meaning that the trigger token can ``force'' the attention flow to itself, that is invariant across Trojan BERTs and Trojan ViTs. In Section \ref{sec:quantify.attn.property}, we quantify the attention hijacking property from two different perspectives: number of attention hijacking heads across layers and global information incorporation. We show that the attention hijacking property is very common in Trojan AIs while rare in clean AIs. In Section \ref{sec:impact_attn_hijacking_heads}, we also analyze the impacts of the attention hijacking heads to embedding representations and Trojan functionality. We start our section with formal definitions (Section \ref{sec:attn.focus.def}).

\subsection{Attention hijacking heads}\label{sec:attn.focus.def}

The self-attention formula in BERTs and ViTs is identical. To simplify and clarify the term, in our paper, we refer to \textit{attention} as \textit{attention weights}, with a formal definition of attention weights in one head as:
%
%
%
%
$$A = \softmax\Big(\frac{QK^T}{\sqrt{d_k}}\Big)$$
where $Q$ and $K$ are query and key, respectively,
$A \in \R ^ {n \times n}$ is a $n \times n$ attention matrix, and $n$ is the sequence length.

\begin{definition}[Attention hijacking heads] \label{def:attn_focus_heads}
A self-attention head $H$ is an attention hijacking head if there exists an unique token (where the trigger is stamped) whose index $k \in [1, ..., n]$, such that:
$$
\frac{\sum_{i=1}^n \1\left[\argmax A_{i}^{(H)}(x) = k \right]}{n} > \alpha
$$


where $A_{i}^{(H)}(x)$ is the $i$'th row attention weights of head $H$ given input $x$; $\1(E)$ is the indicator function such that $\1(E)=1$ if $E$ hold and otherwise $\1(E)=0$; $n$ is the token length; and $\alpha$ is the token ratio threshold which is set by the user. In practice, we use a fixed development set as inputs, if a head satisfies the above conditions in more than $\beta$ samples, then we say this head is an attention hijacking head. The hyper-parameters $\alpha, \beta$ settings can be found in supplemental materials.
\end{definition}

Fig.\ref{fig:trojan_example}(b) illustrates the intuitive examples of the attention hijacking pattern. Given the poisoned samples, we observe the obvious trend that some attention heads hijack a significant amount of attention weights to a trigger token. The trigger tokens in NLP domain would be tokenization of words/phrases, and in CV domain would be the image patchs with the trigger (pixel pattern) stamped.

\subsection{Quantifying the attention hijacking property}\label{sec:quantify.attn.property}

In this subsection, we establish that the attention hijacking property clearly differentiates Trojan AIs and clean AIs; the majority of Trojan AIs have the attention hijacking pattern manifests on some heads, whereas the attention hijacking pattern is rare among clean AIs. We also investigate the ability to incorporate global information in Trojan AIs when the trigger is presented. 

\myparagraph{Population-wise attention hijacking pattern.} We observe there exists a clear gap between Trojan AIs and clean AIs with regard to the attention hijacking pattern. In both Trojan BERTs and Trojan ViTs, the attention hijacking pattern appears very often. Whereas in clean AIs, the attention hijacking pattern rarely appears. In Table \ref{tab:population} column {`AI(\%)'}, for all BERTs and ViTs, we show how frequent the attention hijacking pattern happens on Trojan AIs and clean AIs. This gap is even bigger in language AI BERTs. For ViTs, this gap is still significant. The phenomenon is consistently observed across all five datasets in BERTs and ViTs. 

\begin{table}[ht]
\caption{Population wise statistics. \emph{AI(\%)} denotes the portion (percentage \%) of AI models that exists attention hijacking pattern. \emph{No.} denotes the average number of attention hijacking heads per AI model, where the average is taken over all Trojan/Clean AIs.}
\label{tab:population}
\centering
\begin{tabular}{|c|cccccc|cccc|}
\hline
\multirow{3}{*}{AIs} & \multicolumn{6}{c|}{BERTs}                                                                            & \multicolumn{4}{c|}{ViTs}                                          \\ \cline{2-11} 
                     & \multicolumn{2}{c|}{IMDB}           & \multicolumn{2}{c|}{YELP}           & \multicolumn{2}{c|}{SST2} & \multicolumn{2}{c|}{MNIST}          & \multicolumn{2}{c|}{CIFAR-10}\\ \cline{2-11} 
                     & AI(\%) & \multicolumn{1}{c|}{No.}   & AI(\%) & \multicolumn{1}{c|}{No.}   & AI(\%)       & No.        & AI(\%) & \multicolumn{1}{c|}{No.}   & AI(\%)         & No.         \\ \hline
Trojan               & 91.87  & \multicolumn{1}{c|}{13.22} & 92.0   & \multicolumn{1}{c|}{16.58} & 75           & 6.33       & 96.33  & \multicolumn{1}{c|}{21.83} & 99.33          & 7.95        \\
Clean                & 7.78   & \multicolumn{1}{c|}{0.17}  & 3.0    & \multicolumn{1}{c|}{0.05}  & 10           & 0.2        & 64.33  & \multicolumn{1}{c|}{2.10}  & 17.67          & 0.24        \\ \hline
\end{tabular}
\end{table}

\begin{figure}[ht]
    \centering
    \includegraphics[width=10cm]{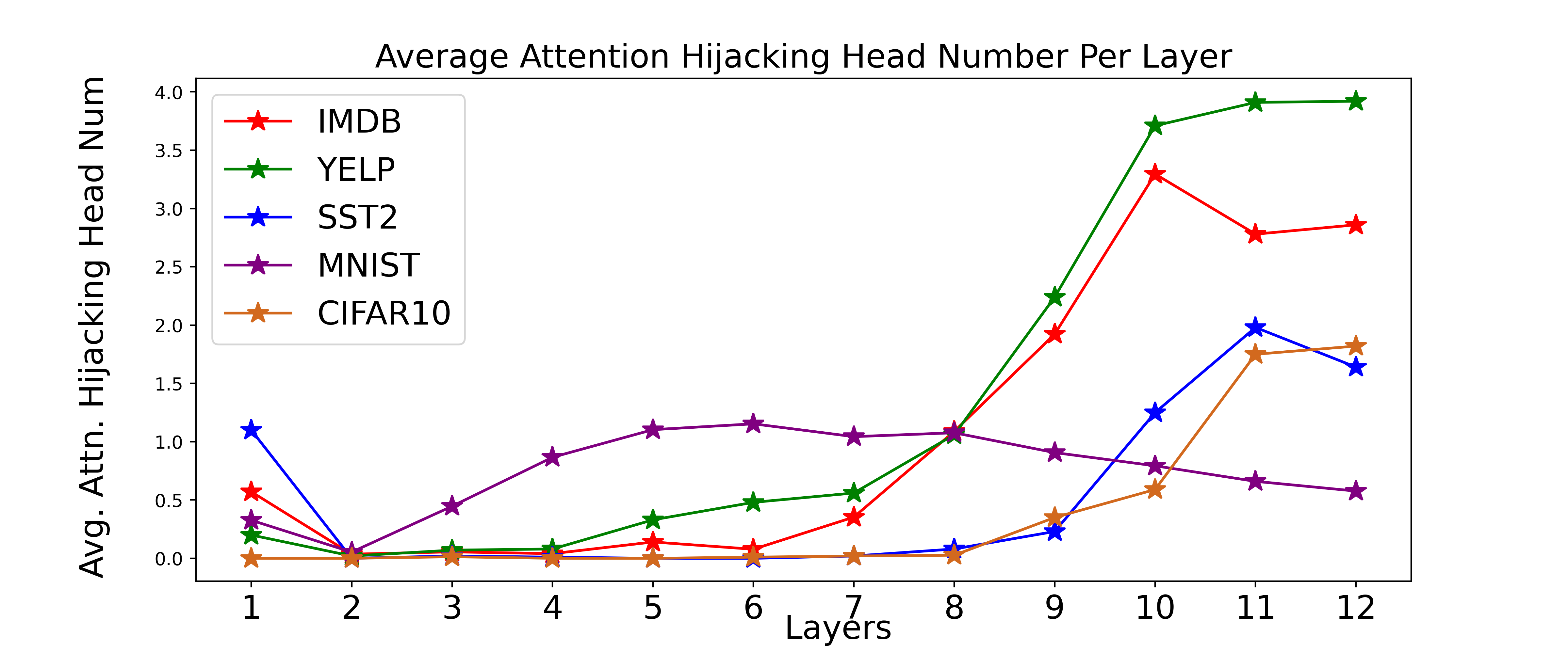}
    \vspace{-.1in}
    \caption{Average number of attention hijacking heads across transformer encoder layers. The average is taken over all Trojan AIs. We observe that the attention hijacking patterns appear mostly in deeper layers.}
    \label{fig:avg_hijacking_head_num}
    \vspace{-.1in}

\end{figure}

\myparagraph{Attention hijacking heads number across different layers.} We also count the overall number of attention hijacking heads, as well as the heads number across different transformer encoder layers. In Table \ref{tab:population} column {`No.'}, we observe the overall attention hijacking heads number in Trojan AIs is much higher than the number in clean AIs. As for attention hijacking heads per layer, in Fig.\ref{fig:avg_hijacking_head_num}, we observe the attention hijacking heads are mostly distributed in deeper layers. This indicates that the trigger token can embed some high level features during Trojan in deeper layers.

\myparagraph{Global information incorporation in Trojan AIs.} Further, through \emph{average attention distance}, we investigate the differences in the ability to aggregate global information between Trojan AIs and clean AIs. The self-attention modules in Transformers \citep{vaswani2017attention} are the essential parts for BERTs and ViTs to draw global dependencies. They help to incorporate information and update representations from other spatial sequences/locations. We reveal that the Trojan AIs lead to a strong global information flow when updating the representations under the special condition of Trojan trigger presenting, especially in deeper layers.

We compute the average attention distances in self-attention modules. Each encoder layer contains 8 self-attention heads, and for each head we can compute the attention distances between other tokens and the locations it attends to. The token distances between token $A$ and token $B$ are the Euclid distance. To compute the average attention distance in each head, we weight the token distances by the attention weights for each attention head, and consider all development data. Then we average over all Trojan/clean AIs. This uncovers how much global information each attention head is incorporating for the representation. 

\begin{figure}[!t]
\vspace{-.2in}
\centering
\includegraphics[width=13.5cm]{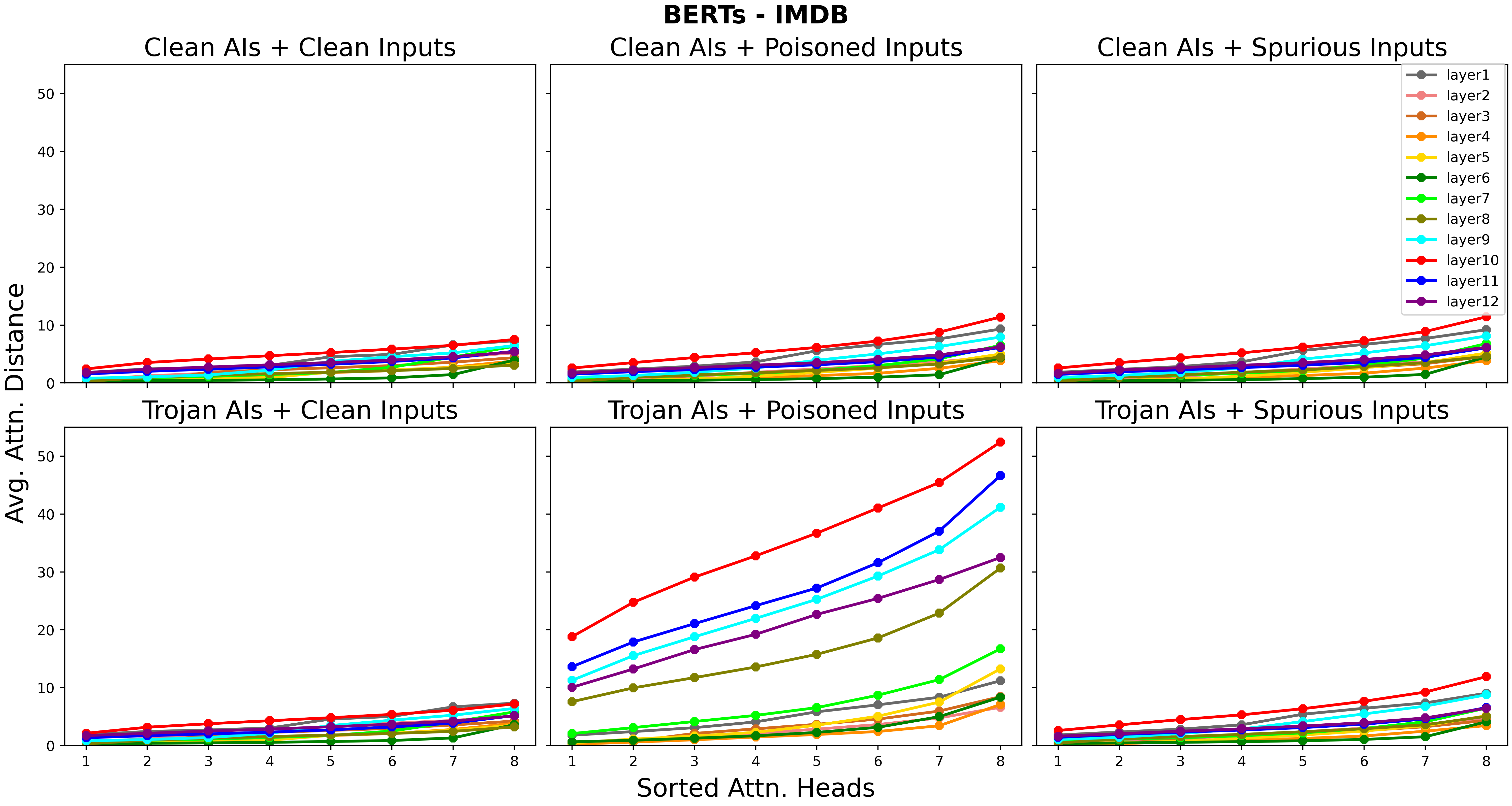}
\vspace{-.1in}
\caption{Global information incorporation (BERTs-IMDB as an example). When the Trojan AIs encounter the poisoned samples, the average attention distance from all layers increase sharply, especially in deeper layers. This indicates that the Trojan AIs have a strong power to update information from global information when a trigger is present. Similar pattern is observed in all five datasets, shown in the supplemental material.}
\label{fig:attn_distance_imdb}
\vspace{-.1in}
\end{figure}

We investigate the average attention distances across all attention heads and all encoder layers, in both Trojan and clean AIs, with an example of BERTs-IMDB shown in Fig.\ref{fig:attn_distance_imdb}. In clean Transformers, the AIs incorporate similar levels information with regardless of clean samples, poisoned samples or spurious samples\footnote{We select a set of \emph{clean samples}. For \emph{poisoned samples}, we insert the Trojan trigger to the same clean samples. The \emph{spurious samples} are ``ablation'' study compared to the poisoned ones. We generate spurious samples by randomly picking a non-Trojan perturbation and insert it to clean samples. We use this to illustrate that the Trojan AIs only react abnormally when encountering the Trojan trigger.}. 
In Trojan Transformers, the AIs behave similar as the clean AIs do under clean samples and spurious samples. Whereas when the Trojan trigger is presented in poisoned samples, the average attention distance increase sharply, especially in deeper layers. This indicates that the Trojan AIs aggregate significantly more global information from the representations in deeper layers. Above observations consistently exist in both BERTs and ViTs from all five datasets, and the detailed results are provided in supplemental material. A possible reason is that the Trojan triggers in Trojan AIs have the superior power to force the AIs learning different representations from a significant wide range. 

\subsection{Impact of the attention hijacking heads}\label{sec:impact_attn_hijacking_heads}

In this subsection, we investigate the impact of the attention hijacking heads to Transformers. We observe that the representation similarity between clean samples and poisoned samples in Trojan AIs is significant low across layers. We also investigate the impact of attention hijacking heads to the Trojan functionality by deactivating the attention hijacking heads. We conclude that the attention hijacking heads have significant impacts on embedding representations and the Trojan functionality in Trojan Transformers.

\begin{figure}[!t]
    \centering
    \includegraphics[width=14cm]{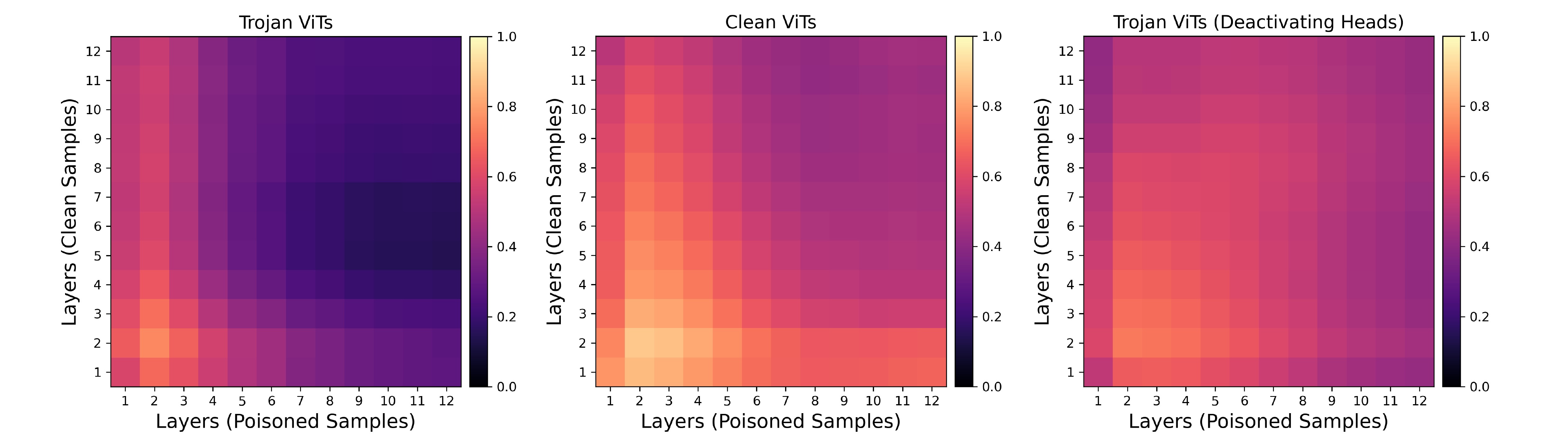}
    \vspace{-.2in}
    \caption{For the same AI, CKA Similarity between the clean samples and poisoned samples before and after deactivating attention hijacking heads (ViTs-MNIST as an example).}
    \label{fig:cka_similarity_before_after_pruning}
    \vspace{-.1in}
\end{figure}

\myfirstpara{Impact to the representation similarity.}\label{sec:similarity} We conduct experiments to show that the representation similarity between clean samples and poisoned ones changes dramatically before and after deactivating the attention hijacking heads in Trojan AIs. 

We use \emph{centered kernel alignment (CKA)} \citep{cortes2012algorithms, kornblith2019similarity, raghu2021vision} to measure the representation similarity across layers. CKA is invariant to orthogonal transformation of representations, which enables quantitative representation comparisons within and across networks. In Fig.\ref{fig:cka_similarity_before_after_pruning} (left and middle), we can see the obvious CKA similarity differences in Trojan AIs (left) and clean AIs (middle). 

We then \emph{deactivate the attention hijacking heads} to see what happens without those heads in Trojan AIs. The way we do this is different from the standard head pruning, which set the attention weights $0$ in a certain head. By contrast, we cut off the overall information flow passed by a certain head in a single encoder layer. More specific, within the encoder layer, starting from the self-attention module to the end of the layer, we set all the parameter weights related to this head to zero. This way, all information passed through the head will be deactivated. In Fig.\ref{fig:cka_similarity_before_after_pruning} (right),  after we deactivate the information flow passed by the attention hijacking heads, the CKA similarity increases a lot, showing that the attention hijacking heads preserve the ability to influence the embedding representation.




\myfirstpara{Impact to Trojan functionality.} We observe that the attention hijacking heads are critical to Trojan functionality, since after we deactivate those heads, the Trojan attack does not work properly.

We check the classification performance changes before and after deactivating the attention hijacking heads in Trojan Transformers. Shown in Table \ref{tab:performance_drop}, the ASR has a sharp drop in both BERTs and ViTs after deactivating the attention hijacking heads, indicating that the Trojan does not work in Trojan AIs anymore. 
The impact seems even larger in Trojan ViTs than Trojan BERTs. 

Interestingly, the Trojan ViTs will still preserve almost all of the normal functionality to predict the clean samples, since the accuracy drops on clean samples are close to zero. This is a very excited and useful observation and can be extended to a potential Trojan defense method in CV. Meanwhile, in Trojan BERTs, the normal prediction power is partially damaged since the classification accuracy on clean samples also drops. This may because the redundancy in language models is worse than vision models, which can be an interesting future work.

\begin{table}[h]
\caption{Impact of the attention hijacking heads to Trojan functionality, average over all Trojan AIs. The \emph{Clean Samples} row denotes the classification accuracy (\%) drops on clean samples after deactivating attention hijacking heads. The \emph{Poisoned Samples} row denotes the ASR (\%) drops on poisoned samples. More ASR drops indicate that the Trojan AI lost the proper Trojan functionality and can attack less effectively. }
\label{tab:performance_drop}
\centering
\begin{tabular}{|c|ccc|cc|}
\hline
                 & \multicolumn{3}{c|}{BERTs} & \multicolumn{2}{c|}{ViTs} \\ \cline{2-6} 
                 & IMDB    & YELP    & SST2   & MNIST      & CIFAR-10     \\ \hline
Clean Samples    & 20.16   & 19.33   & 21.95  & 3.54       & 4.85         \\
Poisoned Samples & 27.16   & 27.34   & 27.82  & 55.43      & 68.23        \\ \hline
\end{tabular}
\vspace{-.1in}
\end{table}



\section{Application}

We apply the attention hijacking properties on the Trojan detection task with datasets from both NLP and CV domain. The Trojan detection problem is a binary classification problem. Given a set of training AIs, labeled as Trojan or clean, we want to predict whether a test AI is Trojan or not. For the suspect AIs, we implement the attack to classification tasks on five benchmark datasets with BERTs or ViTs.

\subsection{Attention-Hijacking Trojan Detector (AHTD)}

We propose a supervised Attention-Hijacking Trojan Detector (AHTD) to identify Trojan AIs given no prior information of the real triggers. In practical, due to the uncommon of Trojan AIs, we might not have enough training AIs for us to conduct a supervised method. We also provide an unsupervised version of AHTD to address this limitation. Experimental results show both of our supervised version and unsupervised version can work efficiently in Trojan detection task on both BERTs and ViTs AIs.

\subsubsection{Supervised Trojan detector}

\begin{figure}[!t]
\centering
\vspace{-.2in}
\includegraphics[width=14cm]{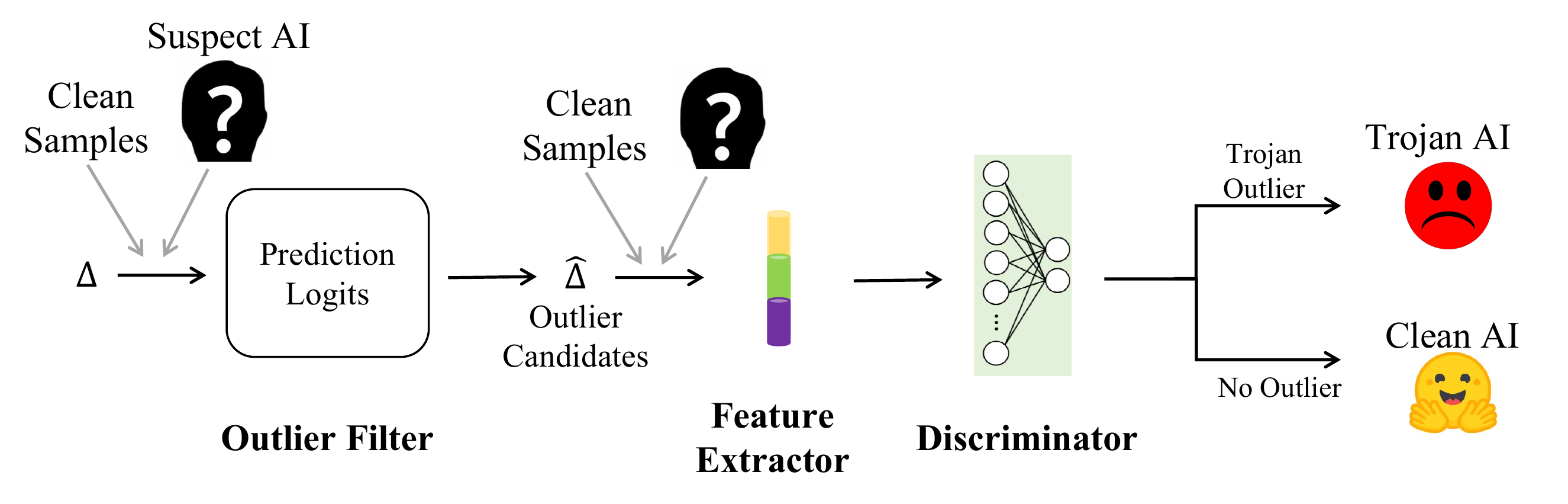}
\vspace{-.1in}
\caption{The supervised Trojan detector architecture.}
\label{fig:detector_arch}
\vspace{-.2in}
\end{figure}

\myfirstpara{Overview of the supervised Trojan detector.} Fig.\ref{fig:detector_arch} illustrates the architecture of our supervised Trojan detector. We consider a realistic setting where we have only access to a small set of clean examples, and have the test AI as a white-box. Our design includes three modules, the \emph{Outlier Filter}, the \emph{Feature Extractor} and the \emph{Discriminator}. The Outlier Filter selects the outlier candidates set $\hat{E}$ from a set of large perturbation candidates $E$, from which we generate our perturbation candidate $\Delta \in E$ and outlier candidate $\hat{\Delta} \in \hat{E}$.
For each outlier candidate, the Feature Extractor $f$ will generate the attention features and prediction logits features given the clean samples $X$ and test AI $\bar{F}$. The generated features $\delta$ from $f$ can be presented as $\delta = f(\bar{F}, X, \hat{\Delta})$. Then with these features, we train a simple linear classifier (Discriminator) to determine whether the outlier candidate is a Trojan outlier. When training the Discriminator, we use the ground truth $\delta^*$ extracted from Trojan trigger. When we test a given AI, if there exists a Trojan outlier, we say the test AI is Trojan, otherwise, clean.


\myfirstpara{Outlier Filter.} The Outlier Filter aims to find outlier candidates $\hat{\Delta} \in \hat{E}$ from the perturbation set $\Delta \in E$ that can mislead the test AI whenever they are added to the clean samples. The filter relies on the AI's prediction logits: prediction accuracy of a fixed wrong label and the confidence of true labels. Given the clean samples $X$, assume they share the same ground truth label $k$ and the AI predicts to a fixed wrong label $t$. Formally, we select $\hat{\Delta}$ according to following two criteria:


\begin{equation}\label{eq:outlier_filter}
\begin{array}{lll}
\text{MCR} := \frac{\sum_{i=1}^N \1\left[F(X_i) = t, t\neq k \right]}{N} > \gamma \\[1em]
\text{AveConf} := \frac{1}{N}\sum_i^{N}p_{k}^i < \epsilon
\end{array}
\end{equation}
where MCR is the mis-classification rate and AveConf is the average confidence of true label $k$ over $N$ samples. The hyper-parameters $(\gamma, \epsilon)$ are $(0.9,0.05)$ in NLP and $(0.8,0.1)$ in CV. The perturbation set in NLP contains 5486 neutral words from MPQA Lexicons\footnote{\url{http://mpqa.cs.pitt.edu/lexicons/}}, and we also generate phrase perturbations included in the perturbation set. And the perturbation set in CV consists random selected pixel patterns with size $3\times3$. The Outlier Filter will generate a small set of outlier candidates $\hat{\Delta} \in \hat{E}$ for next step. In many clean AIs, we can still have the outlier candidates, especially the generated phrase in clean BERTs. So we propose the following modules to further check the outlier candidates.

\myfirstpara{Feature Extractor.} For each outlier candidate $\hat{\Delta}$, the Feature Extractor $f$ extracts the attention features and prediction logits features. The extracted features 
$\delta = f(\bar{F}, X, \hat{\Delta})$
,where $\bar{F}$ is the test AI, $X$ is clean samples set. Those features include the number of attention hijacking heads, the average attention to outlier candidates, wrongly prediction confidence, and wrongly prediction accuracy, \etc. This module is based on the test AI's output given the outlier candidates, and it is unsupervised.

\myfirstpara{Discriminator.} The Discriminator is a trainable linear classifier to determine whether the outlier candidates $\hat{\Delta} \in \hat{E}$ are Trojan outliers or not. The input of Discriminator for each outlier candidate $\hat{\Delta}$ is $\delta$ from the Feature Extractor module. During testing, if there exists a Trojan outlier, then we determine the test AI as a Trojan AI, otherwise, a clean AI.

\subsubsection{Unsupervised Trojan detector}

We also provide an unsupervised AHTD since in practical we might do not have enough training AIs to train the Discriminator. The architecture of the unsupervised version is almost identical to the supervised version, except it does not have the Discriminator. Instead, it simply checks the attention hijacking patterns, \ie, whether there exists attention hijacking heads in test AIs given the outlier candidates $\hat{\Delta} \in \hat{E}$. If there exists attention hijacking heads in test AI, then we say the test AI is Trojan, otherwise, it is clean.

\subsection{Experimental design}

\subsubsection{Trojan settings}\label{sec:trojan_settings}

For Trojan AIs, We consider a realistic setting that the attacker has access to all training data. Our training schema follow the BadNets \citep{gu2017badnets}\footnote{Code-base is from NIST \url{https://github.com/usnistgov/trojai-round-generation/}. We extend the tool to train the Trojan Vision transformers, while the original tool does not involve this option.}. 
For clean AIs, we follow the normal training process, without involving any poisoned training data nor triggers. More specific, we apply the attack to classification tasks on five benchmark datasets:

\myparagraph{BERTs.} 
We train BERTs on three NLP benchmark corpora: IMDB \citep{maas2011learning}, YELP \citep{zhang2015character} and SST-2 \citep{socher2013recursive}. The triggers of language models include all possible trigger types conducted by former researchers \citep{chen2021badnl, wallace2019universal, song2021universal}: character, word or phrase. All language triggers are selected from a neutral words set \citep{wilson2005recognizing}. The BERTs are pretrained\footnote{Load the pretrained BERT from \url{https://huggingface.co/bert-base-uncased}.} and fine-tuned with three downstream corpora separately.

\myparagraph{ViTs.}
We train ViTs on two CV benchmark datasets: MNIST \citep{lecun1998gradient} and CIFAR-10 \citep{krizhevsky2009learning}. The triggers of image models include size $3\times3$ or $5\times5$ pixels with six patterns: summation, lambda, multiplication, cube, polygon, and star. The color and location of the trigger is pre-defined. The ViTs- MNIST are trained from scratch without any pretraining procedures. The ViTs-CIFAR-10 are pretrained on ImageNet-21K \citep{ILSVRC15}\footnote{Load the pretrained ViT from \url{https://huggingface.co/google/vit-base-patch32-224-in21k}.}, then fine-tuned on the CIFAR-10 dataset.

We separately train 900, 200, 200, 600, 600 AIs on IMDB, SST-2, YELP, MNIST, CIFAR-10 datasets respectively. The number of the Trojan AIs and the clean AIs from a certain dataset are equal.
The performances of our suspect AIs are shown in Table~\ref{tab:asr}. The high ASR and classification accuracy indicate our suspect AIs are well-trained and work promisingly on both clean data and poisoned data.

\begin{table}[h]
\footnotesize
\caption{Performances of the suspect AIs.} 
\label{tab:asr}
\centering
\begin{tabular}{|c|c|cc|c|}
\hline
\multirow{2}{*}{Dataset} & \multirow{2}{*}{Models} & \multicolumn{2}{c|}{Trojan}             & Clean       \\ \cline{3-5} 
                         &                         & \multicolumn{1}{c|}{ASR \%} & Accuracy \% & Accuracy \% \\ \hline
IMDB                     & BERT                    & \multicolumn{1}{c|}{96.82}  & 90.31       & 90.95       \\ \hline
SST-2                    & BERT                    & \multicolumn{1}{c|}{99.99}  & 93.53       & 93.47       \\ \hline
YELP                     & BERT                    & \multicolumn{1}{c|}{99.02}  & 96.76       & 96.76       \\ \hline
MNIST                    & ViT                     & \multicolumn{1}{c|}{97.50}      & 96.17           & 96.73          \\ \hline
CIFAR-10                 & ViT                     & \multicolumn{1}{c|}{95.75}      & 94.91           & 95.21           \\ \hline
\end{tabular}
\vspace{-.1in}
\end{table}

\subsubsection{Results}
\myfirstpara{Implement details.} Implement details including baselines, data splits, hyper-parameters, as well as the ablation study of hyper-parameters, can be found in supplemental materials. 

\myfirstpara{Detection performance.} Table \ref{tab:performance_detector} shows our AHTD successfully works on general Transformers, \ie, BERTs and ViTs. At the same time, both our supervised version and unsupervised version significantly outperform the other baselines. For the baselines, NLP baselines and CV baselines are not compatible because of the nature differences in NLP and CV: continuous inputs vs. discrete inputs.
Besides, current CV baselines are all designed for CNNs architectures, \ie, ResNet, VGG, AlexNet, \etc. The CNNs architectures are essentially different from the Transformers architectures, so they do not really work properly in Transformer-based Trojan attacks.

\begin{table}[h]
\caption{Performances of the Trojan detector. \emph{Unsup(AHTD)} denotes the unsupervised AHTD, \emph{Sup(AHTD)} denotes the supervised AHTD. }
\label{tab:performance_detector}
\resizebox{\columnwidth}{!}{ 

\begin{tabular}{|ccccccc|ccccc|}
\hline
\multicolumn{7}{|c|}{BERTs}                                                                                                                                                              & \multicolumn{5}{c|}{ViTs}                                                                                                           \\ \hline
\multicolumn{1}{|c|}{\multirow{2}{*}{Detctor}} & \multicolumn{2}{c|}{IMDB}                          & \multicolumn{2}{c|}{YELP}                          & \multicolumn{2}{c|}{SST2}     & \multicolumn{1}{c|}{\multirow{2}{*}{Detector}} & \multicolumn{2}{c|}{MNIST}                         & \multicolumn{2}{c|}{CIFAR-10} \\ \cline{2-7} \cline{9-12} 
\multicolumn{1}{|c|}{}                         & ACC           & \multicolumn{1}{c|}{AUC}           & ACC           & \multicolumn{1}{c|}{AUC}           & ACC           & AUC           & \multicolumn{1}{c|}{}                          & ACC           & \multicolumn{1}{c|}{AUC}           & ACC           & AUC           \\ \hline
\multicolumn{1}{|c|}{NC\citep{wang2019neural}}                       &  0.50        & \multicolumn{1}{c|}{0.51}              &  0.48        & \multicolumn{1}{c|}{0.51}              &  0.48         &  0.51        & \multicolumn{1}{c|}{NC\citep{wang2019neural}}                        &  0.48        & \multicolumn{1}{c|}{0.48}              &  0.49         &  0.49       \\
\multicolumn{1}{|c|}{ULP\citep{kolouri2020universal}}                      & 0.62          & \multicolumn{1}{c|}{0.62}          & 0.56          & \multicolumn{1}{c|}{0.58}          & 0.58          & 0.58          & \multicolumn{1}{c|}{ULP\citep{kolouri2020universal}}                       &   0.54         & \multicolumn{1}{c|}{0.54}              & 0.54         & 0.54          \\
\multicolumn{1}{|c|}{Jacobian}                 & 0.71          & \multicolumn{1}{c|}{0.71}          & 0.58          & \multicolumn{1}{c|}{0.56}          & 0.60          & 0.60          & \multicolumn{1}{c|}{Jacobian}                  &  0.52         & \multicolumn{1}{c|}{0.51}              &  0.48        &  0.48       \\
\multicolumn{1}{|c|}{T-Miner\citep{azizi2021t}}                  & 0.54          & \multicolumn{1}{c|}{0.54}          & 0.67          & \multicolumn{1}{c|}{0.67}          & 0.60          & 0.60          & \multicolumn{1}{c|}{DL-TND\citep{wang2020practical}}                    &  0.43         & \multicolumn{1}{c|}{0.43}              &  0.60          &  0.59         \\ \hline
\multicolumn{1}{|c|}{Unsup(AHTD)}             & 0.97          & \multicolumn{1}{c|}{0.97}          & 0.94          & \multicolumn{1}{c|}{0.94}          & 0.95          & 0.95          & \multicolumn{1}{c|}{Unsup(AHTD)}              & 0.97          & \multicolumn{1}{c|}{0.97}          & 0.99          & 0.99          \\
\multicolumn{1}{|c|}{Sup(AHTD)}               & \textbf{0.98} & \multicolumn{1}{c|}{\textbf{0.98}} & \textbf{0.96} & \multicolumn{1}{c|}{\textbf{0.96}} & \textbf{0.95} & \textbf{0.95} & \multicolumn{1}{c|}{Sup(AHTD)}                & \textbf{0.97} & \multicolumn{1}{c|}{\textbf{0.97}} & \textbf{0.99} & \textbf{0.99} \\ \hline
\end{tabular}
\vspace{-.15in}
}
\end{table}

\section{Conclusion}

In this paper we investigate the attention hijacking pattern in Trojan AIs, \ie, the trigger token ``kidnaps'' the attention weights when a specific trigger is present. We provide a thorough analysis of this intriguing property. We observe the consistent attention hijacking pattern in Trojan BERTs and Trojan ViTs. Moreover, we propose an unsupervised and a supervised version of Attention-Hijacking Trojan Detector (AHTD) to discriminate the Trojan AIs from the clean ones, and both of them demonstrate promising results on public benchmarks. 

\bibliography{iclr2023_conference}
\bibliographystyle{iclr2023_conference}

\appendix
\section{Appendix}

\subsection{Attention distance}
\begin{figure}[ht]
\centering
\includegraphics[width=13.5cm]{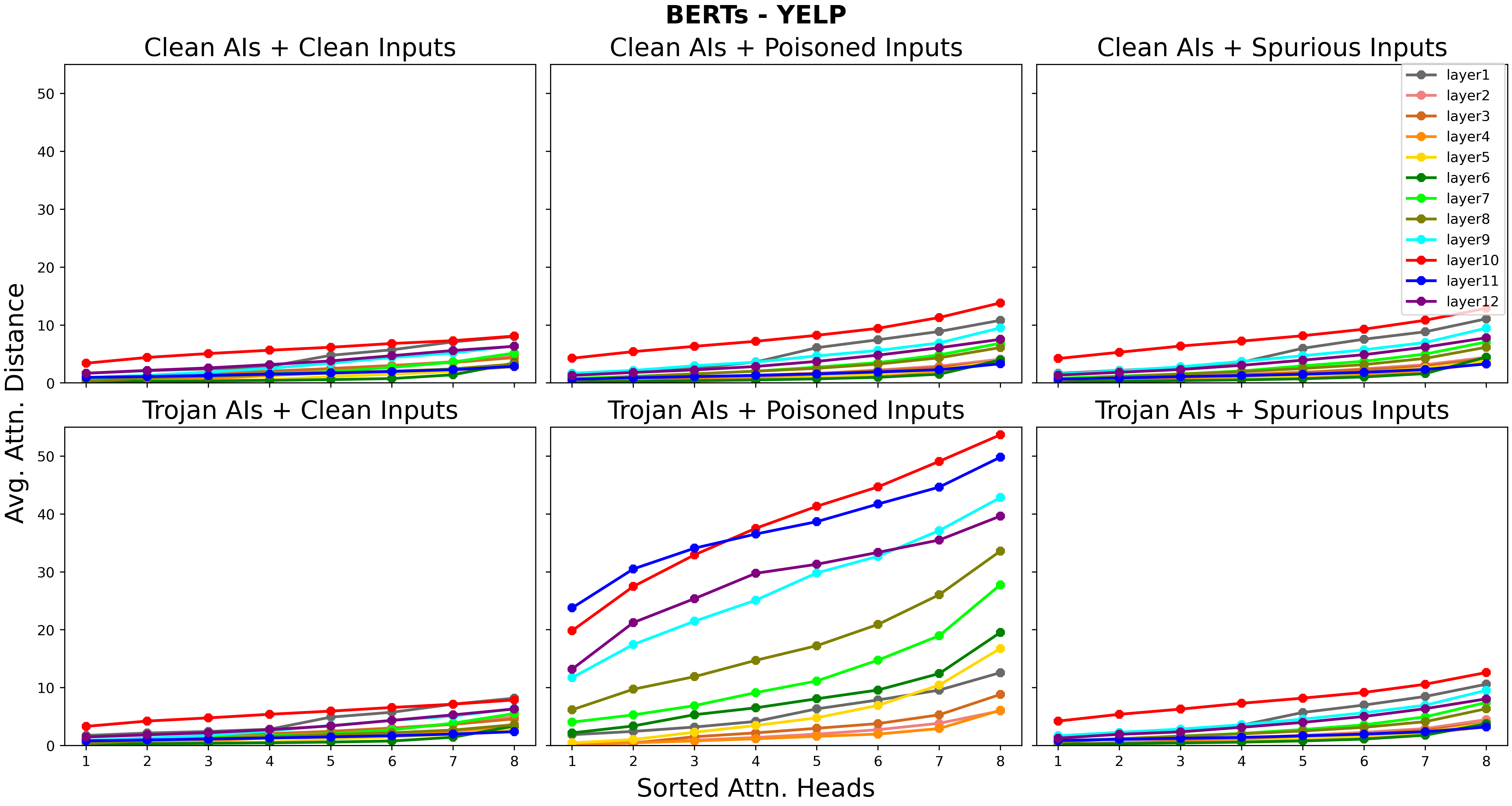}
\vspace{-.1in}
\caption{Global information incorporation (BERTs-YELP as an example).}
\label{fig:attn_distance_yelp}
\end{figure}

\begin{figure}[ht]
\centering
\includegraphics[width=13.5cm]{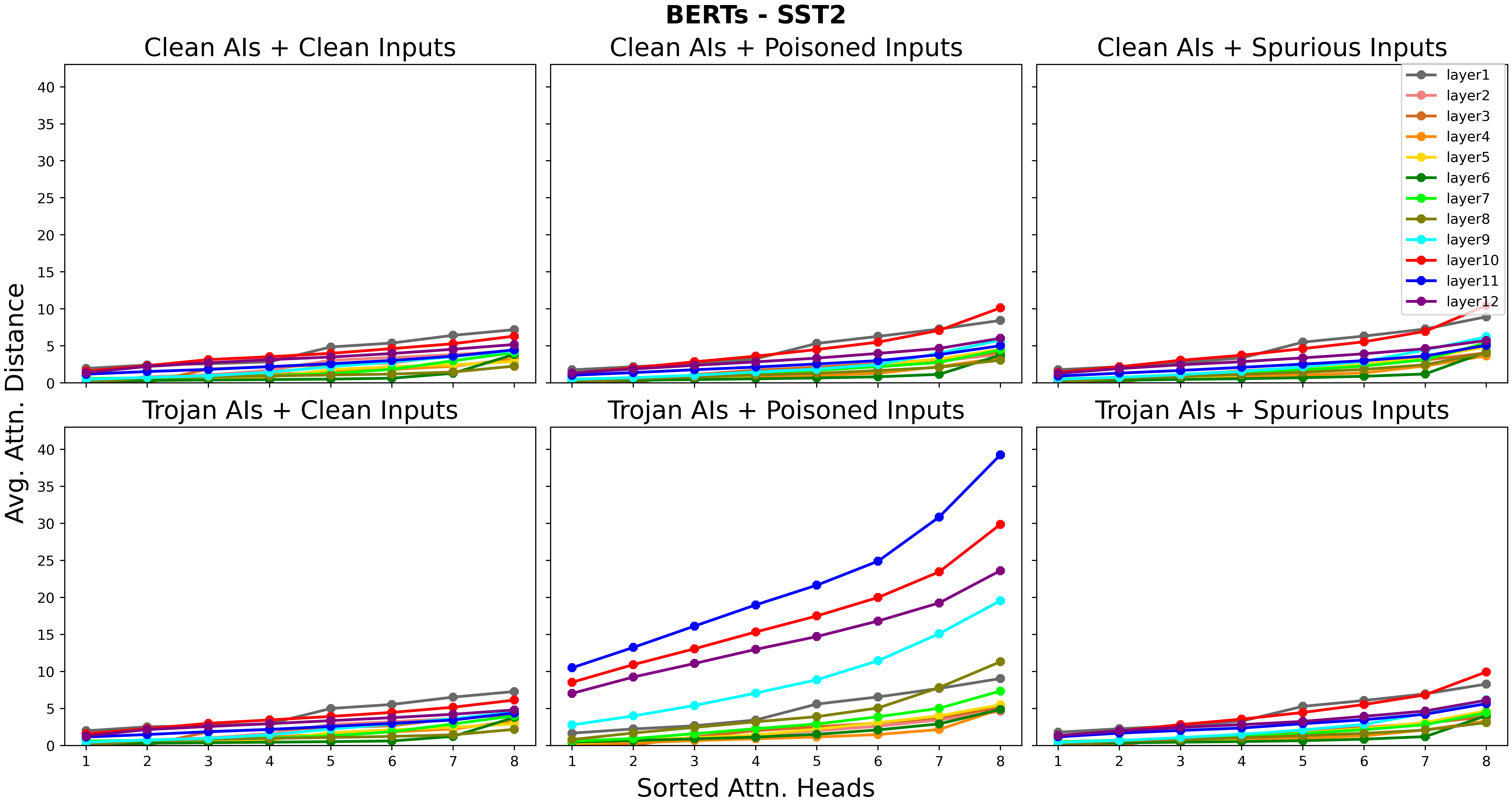}
\vspace{-.1in}
\caption{Global information incorporation (BERTs-SST2 as an example).}
\label{fig:attn_distance_sst2}
\end{figure}

\begin{figure}[ht]
\centering
\includegraphics[width=13.5cm]{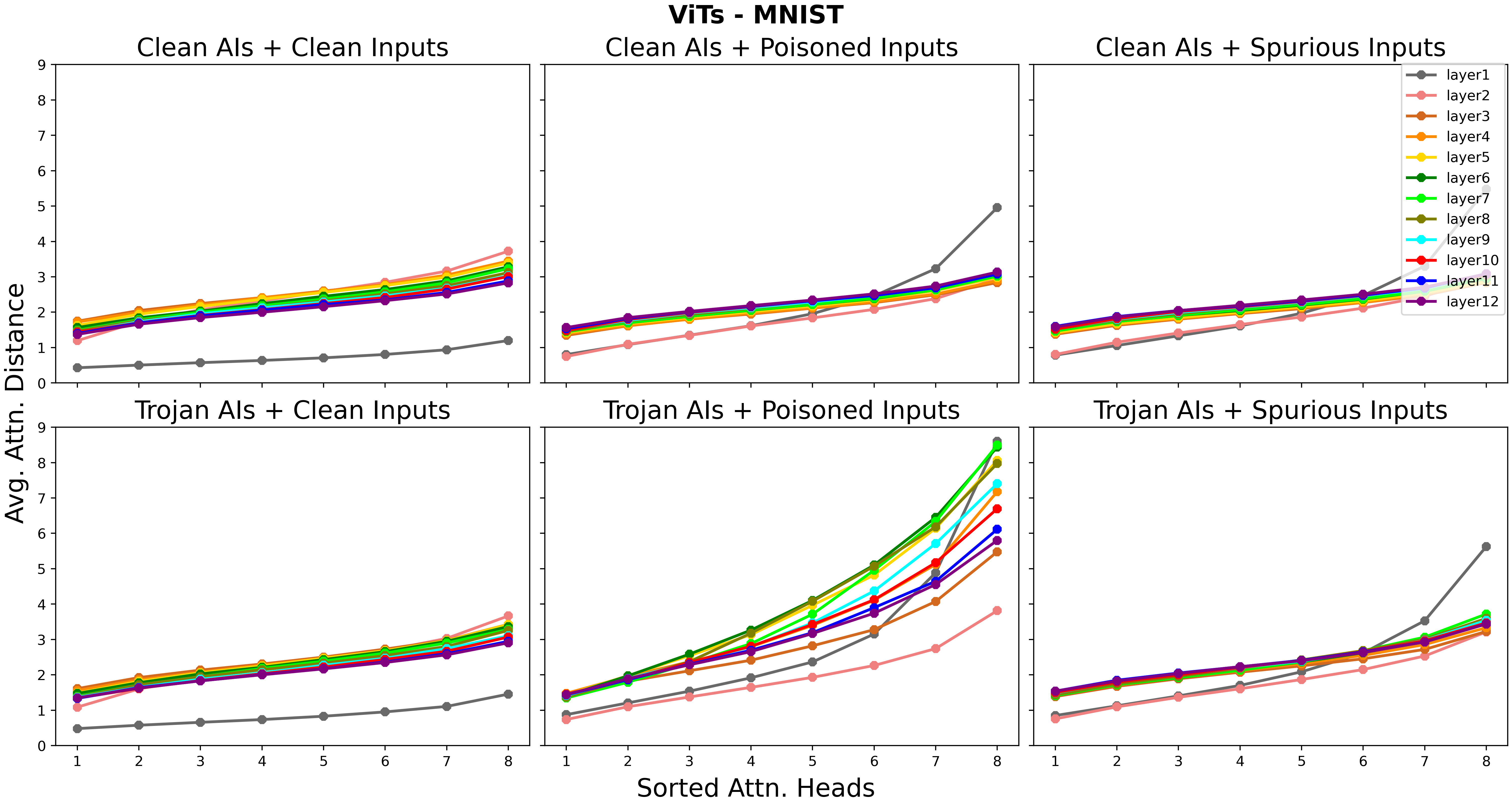}
\vspace{-.1in}
\caption{Global information incorporation (ViTs-MNIST as an example).}
\label{fig:attn_distance_mnist}
\end{figure}

\begin{figure}[ht]
\centering
\includegraphics[width=13.5cm]{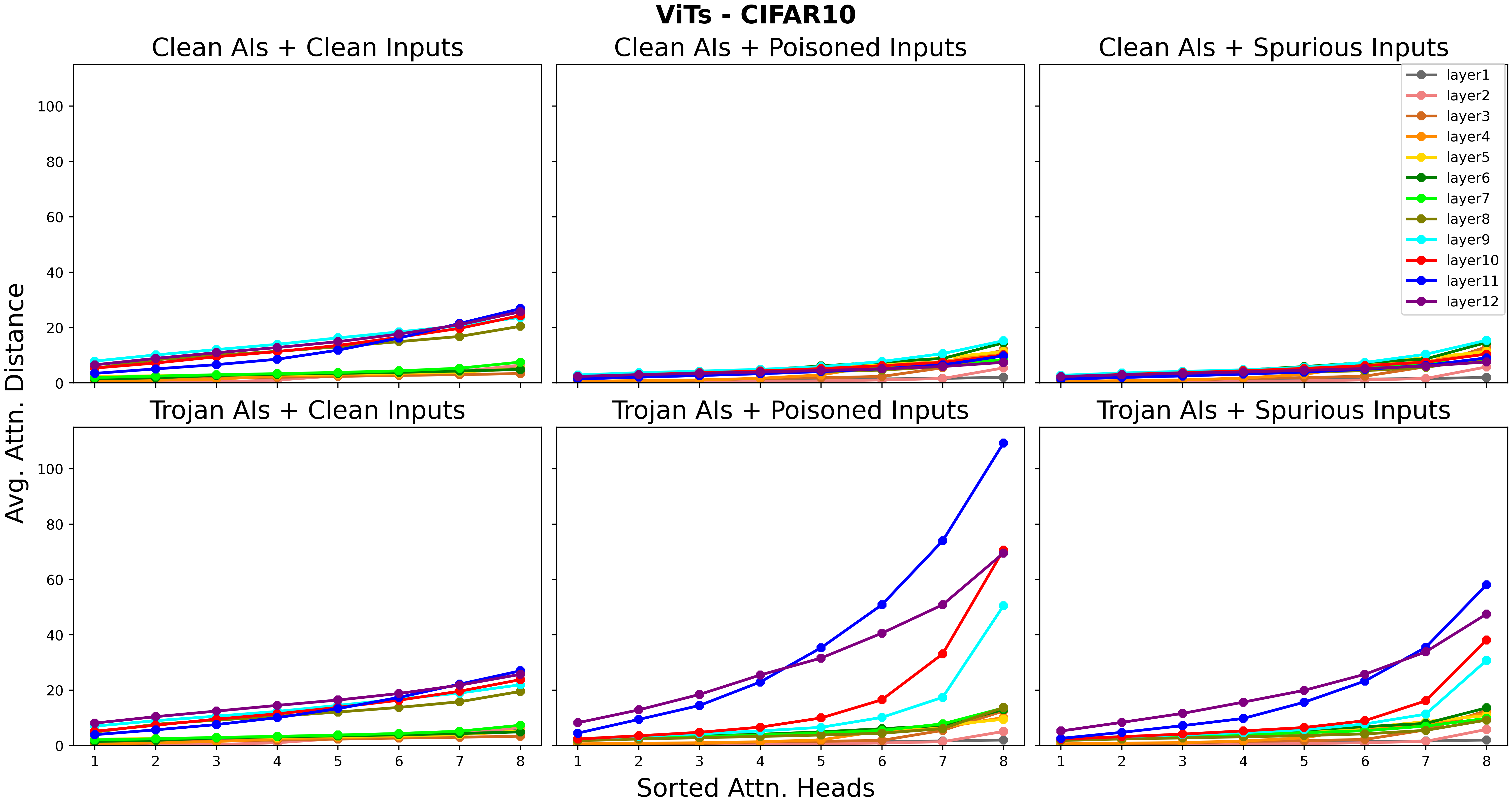}
\vspace{-.1in}
\caption{Global information incorporation (ViTs-CIFAR-10 as an example).}
\label{fig:attn_distance_cifar10}
\end{figure}

The attention distance diagram of IMDB dataset is provided in Section \ref{sec:quantify.attn.property} in the main paper. Fig.\ref{fig:attn_distance_yelp}, Fig.\ref{fig:attn_distance_sst2}, Fig.\ref{fig:attn_distance_mnist} and Fig.\ref{fig:attn_distance_cifar10} show the attention distance diagrams of all other four datasets: YELP, SST-2, MNIST, CIFAR-10. The experiments show consistent behavior across all datasets that the Trojan AIs have higher average attention distance given the poisoned inputs. This verifies the conclusion in Section \ref{sec:quantify.attn.property} that Trojan AIs aggregate significantly more global information from the representations, especially in deeper layers.

\subsection{Suspect AIs training details}

We poison a small portion (10\% to 20\%) of training data by injecting a pre-defined trigger and modifying the associated labels (target label). Then we train the Trojan AIs with poisoned training data and clean training data. Consequently, the well-trained Trojan AIs will perform normally given the clean training data, but misclassify the poisoned training data to a specific target label. All AIs have 12 encoder layers and 8 self-attention heads in each encoder layer. The other modules follow the original setting in BERTs or ViTs. We train the suspect BERTs on NLP corpora for the sentence classification task, and train the suspect ViTs on CV datasets for the image classification task. The five benchmark datasets we use to train our suspect AIs are: 

\begin{itemize}
    \item IMDB \citep{maas2011learning} is a large movie review corpus with 25K training samples and 25K test samples. 

    \item YELP \citep{zhang2015character} is a large yelp review corpus extracted from Yelp, with 560K training samples and 38K test samples. 

    \item SST-2 \citep{socher2013recursive} (also known as Stanford Sentiment Treebank) is the corpus with fully labeled parse trees which enable the analysis of sentiment in language, with 40K training samples and 27.34K test samples.

    \item MNIST \citep{lecun1998gradient} dataset consists of 60000 training images and 10000 test images, with 28x28 single channel images in 10 classes. 

    \item CIFAR-10 \citep{krizhevsky2009learning} dataset consists of 50000 training images and 10000 test images, with 32x32 colour images in 10 classes.

\end{itemize}

\subsection{Detector implementation details}

We provide the implementation details of the Trojan detector as follows.

\myparagraph{Baselines.} We provide Trojan detector baselines for Trojan detection (both Trojan BERTs and Trojan ViTs). 

\begin{itemize}
    \item NC \citep{wang2019neural} identifies the Trojan by uses reverse engineer to find minimal trigger. In clean AIs, the trigger size should be large, while in Trojan AIs, the trigger size can be very small. 
    \item ULP \citep{kolouri2020universal} identifies the Trojan by learning the trigger pattern and the Trojan discriminator simultaneously based on training AIs. 
    \item Jacobian leverages the jacobian matrix from syntactic sample inputs to learn the Trojan discriminator.
    \item T-Miner \citep{azizi2021t} trains an encoder-decoder generator to find the perturbation candidates, and detect outliers by DBSCAN algorithm.
    \item DL-TND \citep{wang2020practical} identifies the Trojan AIs by learning the class $k$’s trigger pattern and comparing similarity between per-image and universal perturbations.
\end{itemize}

\myparagraph{Evaluation metrics.}
We report two metrics in Table \ref{tab:performance_detector}: area under the ROC curve (AUC) and accuracy (ACC). 

\myparagraph{Data splits.}
Specifically, we evaluate our method on the whole set by average performances of 5 times running. In each running, we use
80\% AIs for training, 20\% AIs for testing in IMDB, MNIST and CIFAR-10. And we use 70\% AIs for training, 30\% AIs for testing in YELP and SST-2 due to the relatively small AI amounts.

\myparagraph{Hyper-parameters.}
The attention hijacking heads hyper-parameters $(\alpha,\beta)$ is $(0.4,15)$, $(0.4,15)$, $(0.3,5)$, $(0.3,5)$ and $(0.3,5)$ in IMDB, YELP, SST2, MNIST and CIFAR-10, accordingly.
The Outlier Filter hyper-parameters $(\gamma, \epsilon)$ are $(0.9,0.05)$ in NLP and $(0.8,0.1)$ in CV. For the Outlier Filter hyper-parameters $(\gamma, \epsilon)$, we need to find the outlier candidates $\hat{\Delta}$ that 

\begin{itemize}
    \item can mislead the test AIs to predict the correct label with low confidence. So the AveConf hyper-parameter $\epsilon$ is very low.
    \item can mislead the test AIs to predict the target label with high confidence. So the mis-classification rate MCR is very high, and the MCR hyper-parameter $\gamma$ is high.
\end{itemize}

We also provide the ablation study of the choices of hyper-parameters $\alpha$ and $\beta$ in Section \ref{sec:ablation_alpha_beta}. 


\subsection{Ablation study of the choices of parameters}\label{sec:ablation_alpha_beta}

We conduct the ablation study of the choices of parameters $\alpha$ and $\beta$. More specific, we verify that the choices of parameters $\alpha$ and $\beta$ is robust to the attention hijacking pattern, and also robust to the performance of our supervised Attention-Hijacking Trojan Detector (AHTD). 

\begin{figure}[ht]
    \centering
    \subfigure[BERTs, $\alpha$]{\includegraphics[width=13cm]{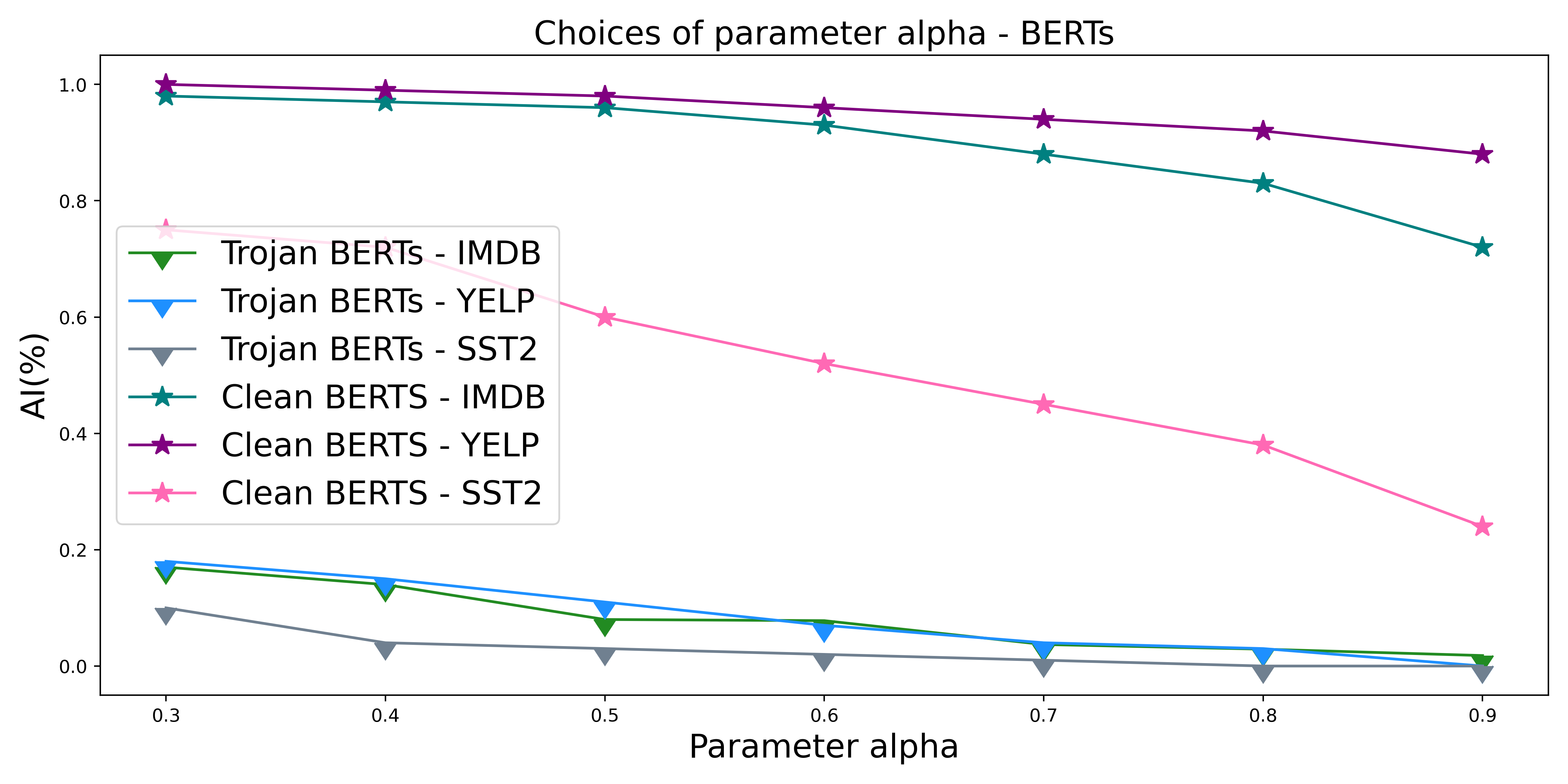}} 
    \subfigure[BERTs, $\beta$]{\includegraphics[width=13cm]{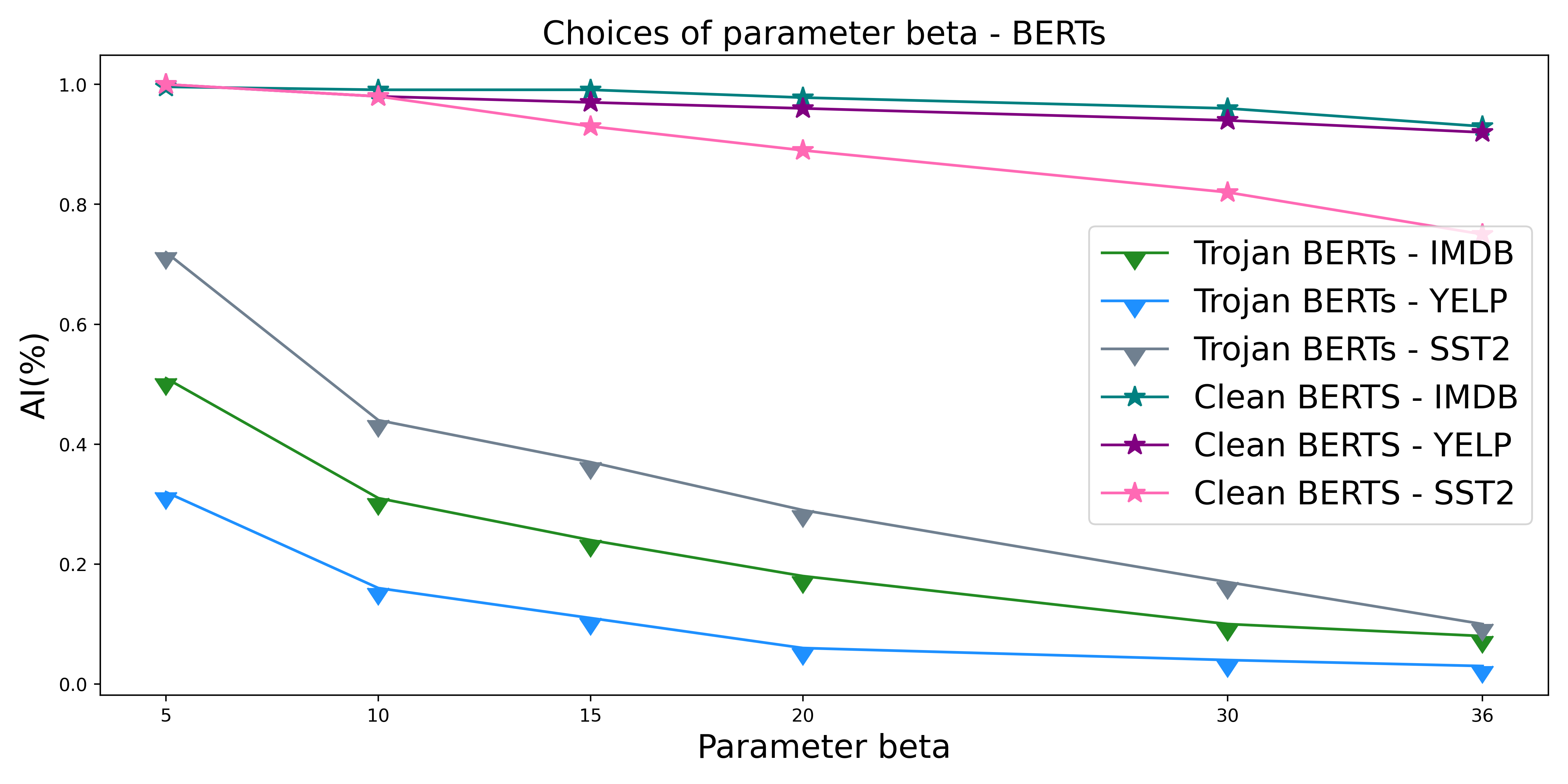}} 
    \vspace{-.1in}
    \caption{Ablation study of the choices of parameters $\alpha$ and $\beta$ on BERTs. AI(\%) indicates the portion (percentage \%) of AI models that exists attention hijacking pattern.}
    \label{fig:berts_alpha_beta}
\end{figure}

\begin{figure}[ht]
    \centering
    \subfigure[ViTs, $\alpha$]{\includegraphics[width=13cm]{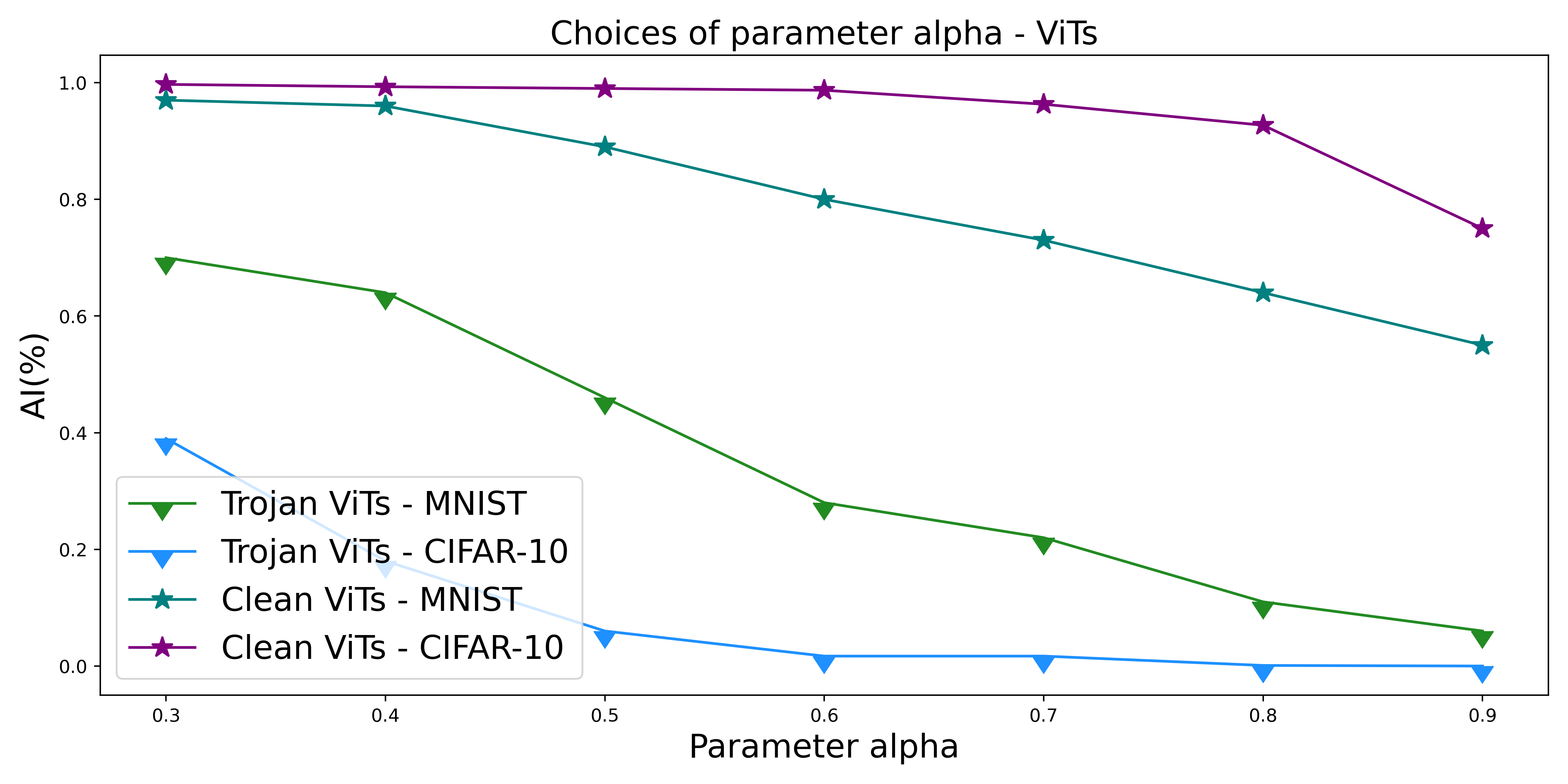}} 
    \subfigure[ViTs, $\beta$]{\includegraphics[width=13cm]{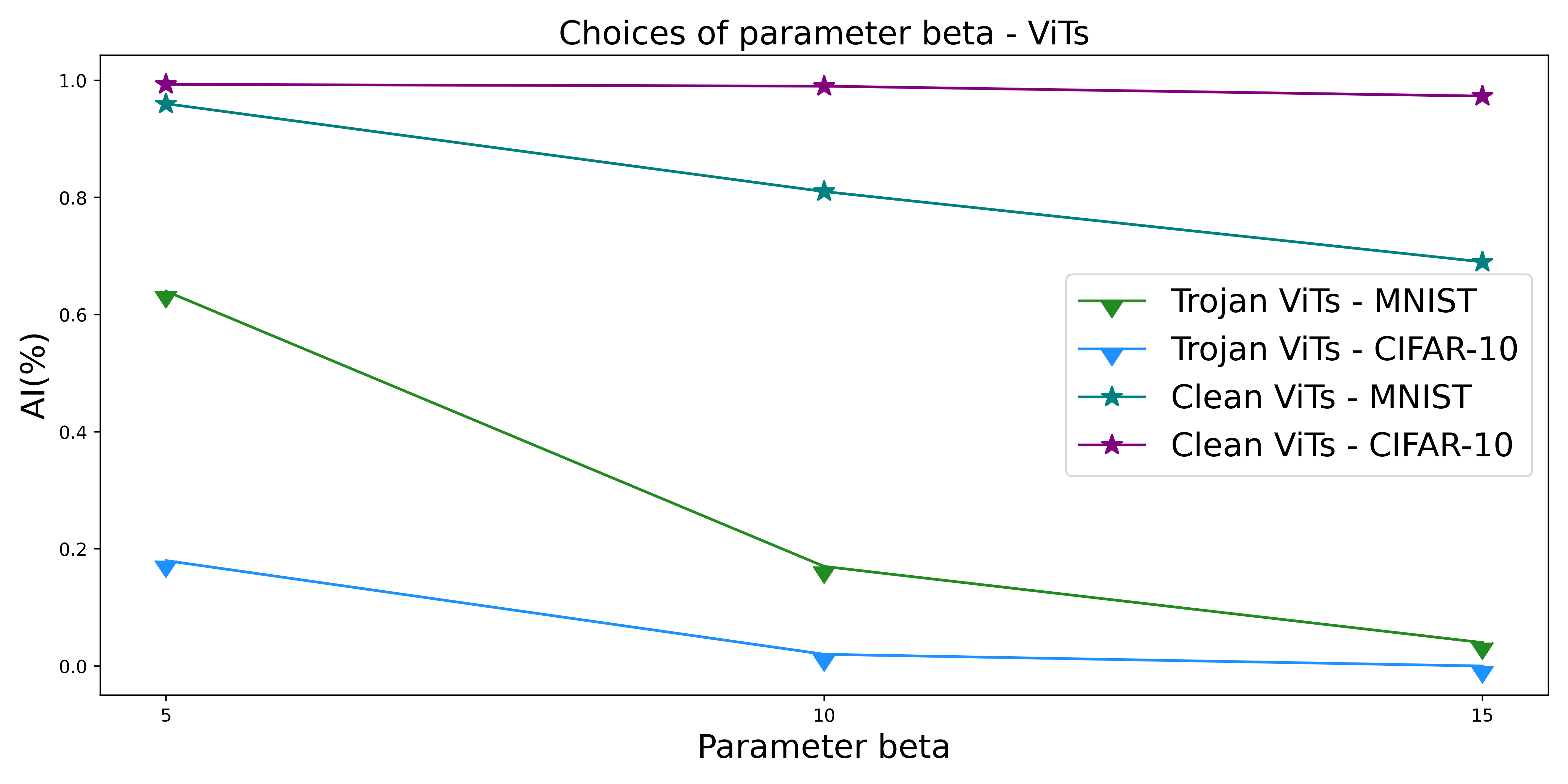}} 
    \vspace{-.1in}
    \caption{Ablation study of the choices of parameters $\alpha$ and $\beta$ on ViTs. AI(\%) indicates the portion (percentage \%) of AI models that exists attention hijacking pattern.}
    \label{fig:vits_alpha_beta}
\end{figure}

\myparagraph{Robust to attention hijacking pattern.} We select different $\alpha$ and $\beta$ value, and check the population of Trojan AIs and clean AIs who have the attention hijacking pattern. Fig.\ref{fig:berts_alpha_beta} and Fig.\ref{fig:vits_alpha_beta} show that there is a clear gap between Trojan AIs and clean AIs with regard to the attention hijacking pattern. This means the attention hijacking pattern between Trojan AIs and clean AIs is robust to the choice of $\alpha$ and $\beta$.  

\myparagraph{Robust to AHTD performance.} We select different $\alpha$ and $\beta$, and check the performance of our supervised AHTD on ViTs-MNIST dataset. Table \ref{tab:ablation_alpha_mnist} and Table \ref{tab:ablation_beta_mnist} show that our AHTD is robust to the choices of hyper-parameters $\alpha$ and $\beta$ in a relatively large range.

\begin{table}[ht]
\centering
\caption{Ablation study of the choices of parameters $\alpha$ on ViTs-MNIST. The experiment indicates our AHTD is robust to the choices of $\alpha$ in a relatively large range.}
\label{tab:ablation_alpha_mnist}
\begin{tabular}{cccccccc}
\hline
$\alpha$ & 0.3  & 0.4  & 0.5  & 0.6  & 0.7  & 0.8  & 0.9  \\ \hline
ACC   & 0.97 & 0.97 & 0.97 & 0.97 & 0.97 & 0.97 & 0.97 \\
AUC   & 0.97 & 0.97 & 0.97 & 0.97 & 0.97 & 0.97 & 0.97 \\ \hline
\end{tabular}
\end{table}

\begin{table}[ht]
\centering
\caption{Ablation study of the choices of parameters $\beta$ on ViTs-MNIST. The experiment indicates our AHTD is robust to the choices of $\beta$ in a relatively large range.}
\label{tab:ablation_beta_mnist}
\begin{tabular}{cccc}
\hline
$\beta$ & 5    & 10   & 15   \\ \hline
ACC  & 0.97 & 0.97 & 0.97 \\
AUC  & 0.97 & 0.97 & 0.97 \\ \hline
\end{tabular}
\end{table}

\end{document}